# Simulating Disengaged Students to Evaluate LLM-based Tutors

**Xianghui MENG, Jionghao LIN***
*The University of Hong Kong, Hong Kong SAR, China*
*jionghao@hku.hk

**Abstract:** Simulated students engineered by computational models can provide a practical way to evaluate tutoring strategies and implement pedagogical approaches used by both human and AI tutors. Such simulations should account for disengaged behaviors, including gaming the system, wheel-spinning, and off-task behavior, because tutors may need to respond differently to these learner states than to engaged students. Our study presents Disengagement-Aware Student Simulators (DAS²), a reproducible pre-deployment protocol that operationalizes five learner-engagement states: engaged, gaming, wheel-spinning, off-task, and mixed, and examines how AI tutor performance varies across these states. For human validation of the learner-state labels, we used ASSISTments09, from which two coders independently labelled 100 sampled tutoring sessions based on anonymized summaries of interaction-log features. They agreed on 84% of sessions (Cohen's κ = 0.78). Among the 84 sessions on which the coders agreed, the human consensus label matched the DAS² rule-based label in 81% of cases (κ = 0.75). Conditioning the simulation on the intended learner state reduced the correctness-rate difference between simulated and authentic sessions from 0.54 to 0.20 for gaming and from 0.51 to 0.18 for wheel-spinning. The fine-tuned Qwen2.5-7B configuration better matched authentic response-time distributions, while the prompt-only GPT-4o configuration produced more distinguishable learner states. We further evaluated five AI tutors from the Claude, Llama, Gemini, Qwen, and GPT model families. Their relative rankings remained stable across learner states and interaction lengths, but absolute performance varied, showing that stable rankings can conceal state-specific differences in tutor support. Human validation also showed that automated tutor evaluation was not fully aligned with human judgment. DAS² provides a pre-deployment evaluation approach for examining how AI tutors respond to different learner-engagement states before deployment.

## 1. Introduction

Simulated students generated by computational models can provide a practical way to evaluate tutoring strategies used by human teachers and AI tutors, as well as to implement pedagogical approaches such as learning by teaching, before they are applied with real students. This is particularly relevant for intelligent tutoring systems whose responses can be examined across different student conditions before classroom deployment. Such simulations should represent not only engaged students but also disengaged behaviors (e.g., gaming, wheel-spinning, and off-task), because tutors may need to respond differently when students guess rapidly, repeatedly struggle with the same skill, or temporarily disengage from the task. Evaluating tutors across these student states can therefore provide a more complete understanding of how they respond to different forms of student behavior.

Prior research provides established behavioral indicators for several forms of disengagement. Gaming the system can involve rapid guessing or repeated requests for final-answer hints (Baker, Corbett, & Koedinger, 2004a; Paquette, de Carvalho, Baker, & Ocumpaugh, 2014). Wheel-spinning refers to repeated attempts on the same skill without reaching mastery (Beck & Gong, 2013), while off-task behavior can be reflected by extended within-session inactivity (Baker, Corbett, Koedinger, & Wagner, 2004b). Recent work continues to develop session-level timing and work-pattern measures from learning-platform logs, showing that behavioral indicators beyond correctness remain relevant for understanding engagement and learning outcomes (Gurung et al., 2025). Recent LLM-based approaches generate tutoring dialogue, condition simulated students on profiles or histories, and manipulate the knowledge level represented by a simulated learner (Macina et al., 2023; Markel, Opferman, Landay, & Piech, 2023; Song, Guo, & Lin, 2026). However, disengagement detectors and LLM-based student simulators are not commonly integrated into the same pre-deployment evaluation protocol.

We address this gap with Disengagement-Aware Student Simulators (DAS²), a reproducible pre-deployment evaluation protocol for AI tutors. operationalizes five session-level learner-engagement states: engaged, gaming, wheel-spinning, off-task, and mixed. It first examines whether these operationalized states can be consistently interpreted by human coders, then uses them to condition simulated students and evaluates how closely the resulting behaviors resemble authentic tutoring logs. Finally, DAS² examines whether tutor performance, measured through relevance, scaffolding, and engagement recovery, as well as tutor rankings, varies across learner states. These states are used as session-level evaluation categories rather than enduring psychological traits or diagnoses, and they do not represent all possible forms of learner engagement. We address three **R**esearch **Q**uestions:

**RQ1**: To what extent do student-state labels generated by the DAS² detector rules align with human judgments of engaged, gaming, wheel-spinning, off-task, and mixed behavior?
**RQ2**: To what extent do simulated students reproduce the behaviors of their intended student-engagement states?
**RQ3**: How does AI tutor performance vary across simulated student-engagement states?

DAS² makes three contributions. *First,* it operationalizes engaged, gaming, wheel-spinning, off-task, and mixed behavior as session-level learner states using established behavioral indicators and examines whether these states can be consistently interpreted by human coders. *Second*, it evaluates both behavioral alignment and learner-state controllability, including response timing, correctness, hint use, disengagement patterns, and state separability, revealing that closer alignment with authentic logs does not necessarily produce more distinguishable learner states. *Third*, it provides a pre-deployment evaluation protocol for comparing AI tutors across learner states, distinguishing changes in absolute tutor performance from changes in tutor rankings and examining the alignment between automated and human evaluation.

## 2. Related Work

### *2.1 Disengagement Detection*

Student interaction logs contain behavioral signals beyond response correctness that can indicate disengagement. Gaming the system has been associated with rapid guessing and excessive hint use (Baker et al., 2004a; Paquette et al., 2014), wheel-spinning with repeated practice on the same skill without mastery (Beck & Gong, 2013), and off-task behavior with extended inactivity during learning (Baker et al., 2004b). More recent work continues to use timing and interaction patterns to characterize engagement from learning-platform logs (Gurung et al., 2025). DAS² builds on these established indicators to operationalize engaged, gaming, wheel-spinning, off-task, and mixed behavior as session-level states, validate their interpretability through human annotation, and use them as conditions for student simulation.

*2.2 Large Language Model Based Student Simulation*

Student simulation has long supported the development and evaluation of intelligent tutoring systems, including cognitive, rule-based, and learned approaches such as the Apprentice Learner architecture (MacLellan et al., 2016). Recent LLM-based systems enable richer student-tutor interactions: GPTeach uses GPT-based simulated students for teacher training (Markel et al., 2023), while MathDial provides simulated tutoring dialogues grounded in mathematical reasoning (Macina et al., 2023). Other work examines whether simulated students can represent particular learner characteristics, including linguistic, behavioral, and cognitive dimensions (Scarlatos et al., 2026), student histories and behavioral tendencies (Duan et al., 2026), and knowledge levels manipulated through model unlearning and relearning (Song et al., 2026). DAS² extends this direction to learner-engagement states and jointly evaluates behavioral alignment and state separability, asking not only whether generated behavior resembles authentic sessions but also whether the intended engagement states remain distinguishable.

*2.3 Evaluating Tutor Performance Across Student States*

Tutor effectiveness may depend on the student behavior encountered: repeated guessing, persistent difficulty, and off-task behavior can require different instructional responses. Evaluating tutors only through aggregate performance can therefore conceal differences across student conditions, consistent with broader evidence that aggregate educational-system measures may obscure heterogeneous outcomes (Doroudi & Brunskill, 2019). Recent tutor-evaluation work further motivates examining performance across student behavioral conditions and comparing models through evaluation-based rankings (Niousha et al., 2026; Macina et al., 2025). These forms of evidence need not agree: rankings may remain stable even when absolute performance changes across student states. DAS² therefore evaluates both state-specific tutor scores and relative tutor rankings, revealing performance differences that may be hidden by a single aggregate score.

## 3. DAS²: Data, Labels, and Simulators

Figure 1 summarizes the DAS² workflow in six stages. **Stage 1, Authentic student interaction logs**, standardizes four public datasets into a common format while assigning them distinct roles in the evaluation. ASSISTments09 serves as the primary dataset for simulator training and human validation; ASSISTments12 is used for related-domain evaluation, Eedi for cross-dataset evaluation, and EdNet-KT1 for response-time and off-task analyses where the required signals are available. **Stage 2, Literature-grounded disengagement detectors**, applies behavioral indicators from prior research to identify gaming the system, wheel-spinning, and off-task behavior. **Stage 3, Student-state assignment**, combines these detector outputs to assign each session to one of five student-engagement states: engaged, gaming, wheel-spinning, off-task, or mixed. These states are session-level evaluation categories rather than enduring student traits. **Stage 4, DAS² simulator configurations**, examines three main approaches with different uses of authentic student data. V1 is a prompt-only GPT-4o simulator instructed to represent a specified student state; V2 extends V1 by providing three similar authentic sessions as examples; and V3 is a LoRA-fine-tuned Qwen2.5-7B simulator trained on ASSISTments09 interaction sequences. An additional V3-cal configuration adjusts the generated correctness distribution through post-processing rather than training a new simulator. **Stage 5, Behavioral alignment and state separability**, compares simulated sessions with authentic sessions to examine both how closely their behavioral distributions align and whether the five intended student states remain distinguishable. **Stage 6, AI Tutor evaluation**, allows simulated students to interact with five AI tutors and examines whether tutor performance and relative rankings vary across student-engagement states. The primary evaluation uses 10-turn interactions, with 20-turn interactions used as a robustness check, and automated tutor ratings are further compared with human judgments.

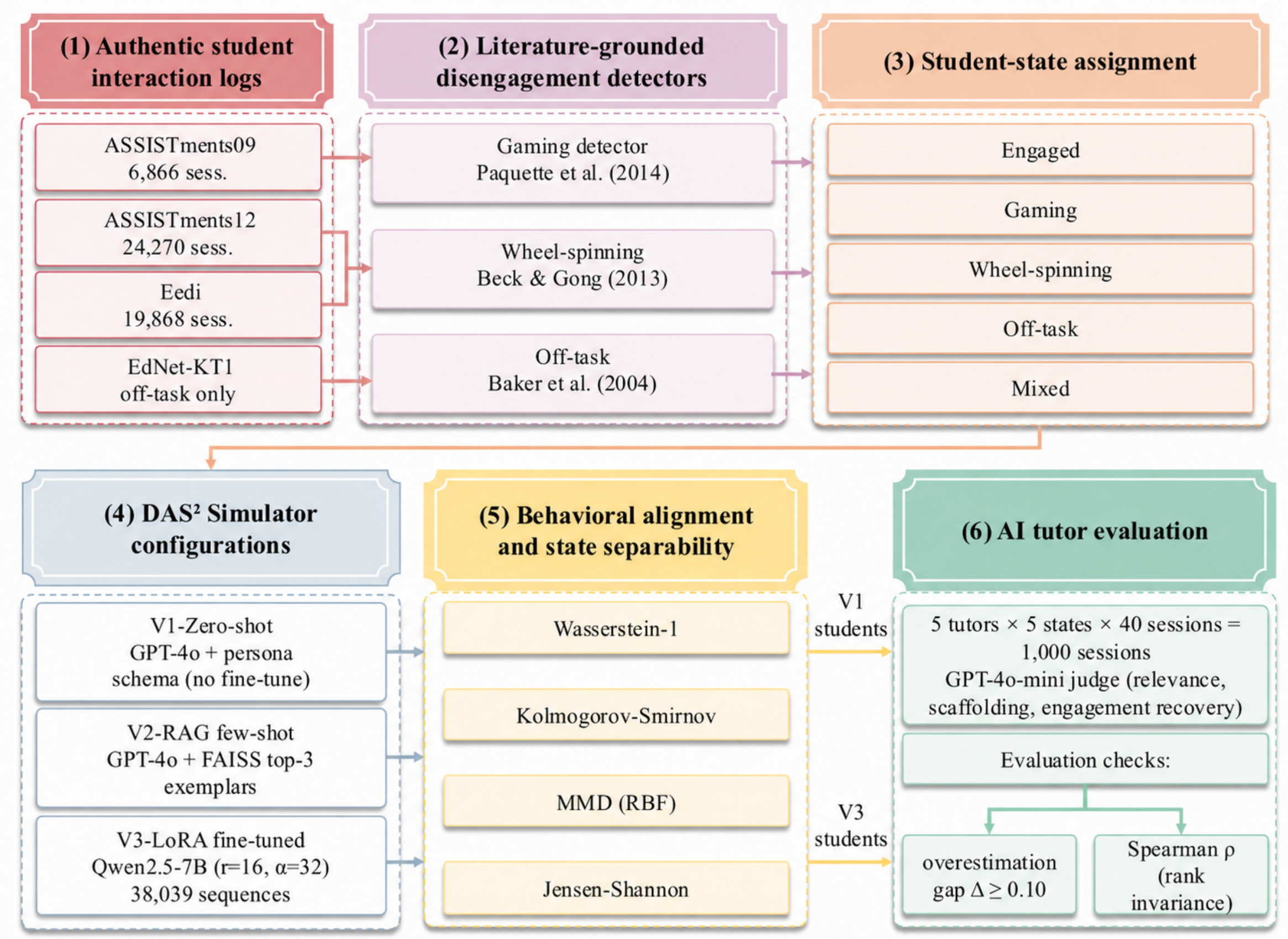

*Figure 1.* The DAS² workflow in six stages: authentic student interaction logs, literature-grounded disengagement detectors, student-state assignment, simulator configurations, behavioral alignment and state separability, and AI tutor evaluation.

To evaluate the different stages of DAS², we use complementary measures rather than a single overall metric. Wasserstein-1, Kolmogorov-Smirnov D, MMD, and Jensen-Shannon divergence assess behavioral alignment between simulated and authentic sessions; silhouette scores and classification accuracy assess student-state separability; and state-specific tutor scores and rank correlations examine whether tutor performance changes across student-engagement states. Human ratings are additionally used to evaluate the validity of the automated tutor evaluation.

### *3.1 Datasets and Detectable Learner States*

Table 1 summarizes the four public datasets used in DAS². ASSISTments09 (Feng, Heffernan, & Koedinger, 2009) serves as the primary dataset for simulator training and human validation of the student-state labels. ASSISTments12 provides a related-domain dataset for examining whether behavioral alignment transfers beyond the primary dataset. Eedi (Wang et al., 2021) provides a structurally different student-response dataset for examining cross-dataset generalization. EdNet-KT1 (Choi et al., 2020) is used only for response-time and off-task analyses because the version used in this study does not provide the correctness information required to identify gaming and wheel-spinning.

### *3.2 Rules for Labeling Learner States*

DAS² applies literature-grounded behavioral rules to assign student-engagement states at the session level. **Gaming** is identified when a student gives an incorrect response in less than 2 seconds, requests a bottom-out hint on the first attempt, or uses at least three hints within five consecutive steps with response times below 30 seconds (Baker et al., 2004a; Paquette et al., 2014). **Wheel-spinning** is identified when a student makes more than 10 attempts on the same knowledge component (KC) without achieving three consecutive correct responses, or makes more than eight attempts with mean correctness below 0.35 (Beck & Gong, 2013). **Off-task behavior** is identified when an interaction follows an inactivity gap longer than 5 minutes but shorter than 30 minutes within the same session (Baker et al., 2004b). Sessions without gaming or wheel-spinning and with less than 10% off-task behavior are classified as **engaged**, while sessions with overlapping or otherwise unresolved behavioral patterns are classified as **mixed**.

*Table 1. Datasets, Scale, and Student-State Labels Used in DAS²*

| Dataset | Sessions | Gaming | WS | Off-Task | Role |
|---|---|---|---|---|---|
| ASSISTments09 | 6,866 | 7.4% | 1.9% | 1.1% | Training and human validation |
| ASSISTments12 | 24,270 | 3.9% | 0.0% | 1.0% | Related-domain evaluation |
| Eedi | 19,868 | 31.2% | 0.0% | 0.1% | Cross-dataset evaluation |
| EdNet-KT1 | 11,208 | n/r | n/r | 3.4% | Response-time and off-task evaluation |

*Note.* WS = wheel-spinning; n/r = not reported. ASSISTments12 and Eedi yielded no wheel-spinning sessions under the implemented rule in the analysed samples. EdNet gaming and wheel-spinning are not reported because correctness is unavailable.

*3.3 Session-Level Personas*

The five student states identified above are used to guide simulated student behavior. Specifically, each simulator receives a student-state persona specifying that it should represent an engaged, gaming, wheel-spinning, off-task, or mixed student session. The student state describes the behavioral category identified in authentic interaction logs, whereas the corresponding persona provides the simulator with instructions for generating behavior representative of that state.

*3.4 Human Validation of Student-State Labels*

To examine whether the DAS² student-state labels were interpretable to humans, we sampled 100 ASSISTments09 sessions, with 25 sessions from each of four DAS²-assigned sampling groups: engaged, gaming, wheel-spinning, and a combined off-task/mixed group. Two coders independently assigned one of the five student-state labels to each session based on anonymized interaction summaries containing correctness patterns, response-time statistics, hint use, number of steps, inactivity gaps, repeated attempts on the same KC, action traces, and number of KCs. The coders were not shown the DAS²-assigned label, student identifiers, or each other's annotations. The annotation materials included one worked example for each student state. Because these examples were drawn from the sampled sessions, they were used only to familiarize coders with the labeling procedure and were not treated as independent validation examples. Disagreements between the two coders were not adjudicated. All 100 sessions were retained when calculating inter-coder agreement, while comparison between human and DAS² labels was restricted to the 84 sessions on which both coders independently assigned the same state.

*3.5 Student Simulator Configurations*

DAS² includes three main student-simulation configurations, **V1**, **V2**, and **V3**, which generate simulated interaction sessions representing a specified student-engagement state. The configurations differ in how authentic student interaction data are incorporated into generation. Their generated sessions are subsequently compared with authentic sessions to evaluate behavioral alignment and whether the intended student states remain distinguishable. **V1** is a prompt-only GPT-4o configuration with no fine-tuning. It receives an instruction specifying the target student state and generates step-level behavior, including correctness, hint use, response time, and action type. **V2** extends V1 by retrieving three similar labelled sessions using an eight-dimensional feature representation and FAISS-based nearest-neighbor search. The retrieved interaction sequences are provided as few-shot examples. Because the retrieval pool does not explicitly exclude held-out evaluation sessions, V2 is treated as exploratory rather than as evidence of held-out generalization. **V3** uses Qwen2.5-7B-Instruct fine-tuned with LoRA on 38,039 ASSISTments09 instruction-response sequences. A single shared adapter is used across student states, while the target student-state persona is specified in the system prompt. **V3-cal** is not a separately trained simulator. Instead, it applies post-processing to V3 outputs by changing surplus correct responses to incorrect responses so that the generated correctness distribution more closely matches the target state, while leaving response time, hint use, interaction order, and persona unchanged.

## 4. Evaluation Protocol

### *4.1 Behavioral Alignment Metrics*

We evaluate how closely simulated sessions resemble authentic student sessions using four complementary distributional measures. Wasserstein-1 (W1) measures the difference between observed and simulated response-time distributions after applying a log transformation; lower values indicate closer alignment. Kolmogorov-Smirnov D (KS D) measures the maximum difference between the corresponding response-time distributions, with lower values indicating greater similarity. Maximum Mean Discrepancy (MMD) provides an additional distributional comparison of log-transformed response times in the primary ASSISTments09 analysis. Jensen-Shannon (JS) divergence compares the joint distributions of correctness, hint use, and response time. For all four measures, smaller values indicate closer alignment between simulated and authentic student behavior. Confidence intervals for the primary W1 analysis are estimated using 1,000 bootstrap samples.

### *4.2 Student-State Representation and Separability*

We next examine whether the student simulator can generate behavior corresponding to the intended student-engagement states. We first conduct a **persona ablation within V3** to isolate the effect of the student-state instruction while keeping the underlying simulator unchanged. **V3-no-persona** receives the engaged-state instruction regardless of the target state, whereas **V3-persona** receives the instruction corresponding to the intended state. **V3-cal** additionally adjusts the generated correctness distribution through post-processing. Comparing V3-no-persona with V3-persona therefore examines whether explicitly specifying the target student state improves alignment with authentic sessions from that state.

We then examine whether the generated student states remain behaviorally distinguishable across simulator configurations. This analysis compares V1, V3-persona, and V3-cal. V2 is excluded from this primary comparison because its retrieval pool does not enforce complete separation from the evaluation data and is therefore treated as exploratory. Silhouette scores measure how clearly sessions from different student states separate in the behavioral feature space, while five-state classification accuracy measures how accurately the intended state can be predicted from the generated behavior. Because five states are evaluated, chance-level classification accuracy is 20%. These analyses distinguish two properties of student simulation: whether the generated behavior resembles authentic sessions from the intended state and whether different intended states remain distinguishable from one another.

### *4.3 AI Tutor Evaluation Across Student States*

We evaluate five AI tutors, Claude-3.5-Haiku, Llama-3.1-70B, Gemini-2.0-Flash, Qwen2.5-72B, and GPT-4o, through interactions with simulated students representing the five engagement states. Each tutor interacts with each student state in 40 sessions. One turn consists of one simulated student action followed by one tutor response. The primary evaluation uses 10-turn interactions, while an independent 20-turn evaluation examines whether the results remain consistent over longer interactions. Tutor responses are primarily evaluated by GPT-4o-mini on three dimensions: relevance, scaffolding, and engagement recovery, each scored from 1 to 5. The three ratings are combined into a 0 to 1 performance score. To examine scoring consistency, 50 transcripts are independently rescored using the same judge configuration. Human ratings are collected to examine how well the automated evaluation corresponds to human judgments. For each tutor, we compare its average score with engaged students against its average score across gaming, wheel-spinning, off-task, and mixed students. A difference of at least 0.10 is treated as meaningful within this evaluation protocol. We also compare tutor rankings under engaged and disengaged conditions using Spearman's correlation to determine whether changes in student state alter the relative ordering of tutors. Finally, 200 transcripts are independently scored by Claude-3.5-Haiku to examine whether the results depend on the choice of automated judge.

## 5. Results

### *5.1 Human Validation of Student-State Labels*

The DAS² student-state labels showed substantial agreement with human judgments. Two coders independently labelled 100 ASSISTments09 sessions and agreed on 84 sessions (84%; Cohen's $\kappa = 0.78$). Among these 84 agreed cases, the human labels matched the DAS²-assigned labels for 68 sessions (81%; $\kappa = 0.75$). These results indicate that the operationalized student states were generally interpretable to human coders. Among the 21 rule-gaming cases in the 84-session agreed subset, 9 were human-labelled gaming, 7 wheel-spinning, 4 mixed, and 1 off-task, so gaming precision =9/21=0.43. This suggests that gaming was particularly difficult to distinguish from repeated unsuccessful practice. Many of these boundary cases involved a single knowledge component, incorrect responses, and hint use, making them behaviorally similar to wheel-spinning. In contrast, all sessions classified by DAS² as engaged, wheel-spinning, or off-task within the agreed subset were assigned the same label by the human coders, corresponding to a precision of 1.00 for these states. However, this should be interpreted cautiously, particularly for off-task behavior, because only five DAS²-classified off-task sessions were available. Overall, the results suggest that the five-state framework is generally interpretable, while the distinction between gaming and wheel-spinning requires further refinement.

### *5.2 Behavioral Alignment and Student-State Distinguishability*

In Table 2, the simulator configurations differed in how closely they reproduced authentic student behavior. On ASSISTments09, the fine-tuned Qwen2.5-7B configuration (V3) showed closer response-time alignment than the prompt-only GPT-4o configuration (V1), with W1 decreasing from 1.00 to 0.50. Accompanying reductions in KS (0.60 to 0.32), MMD (0.31 to 0.10), and JS (0.62 to 0.30) show that the improvement is not limited to a single statistic. Exact-equality tests yield $p<0.001$, which rejects identity but does not establish practical magnitude. Table 2 summarizes cross-dataset response-time alignment and the persona-conditioning ablation. For engaged sessions, the V3-versus-V1 ordering transfers to ASSISTments12 (0.49 vs. 0.94), while Eedi shows substantial domain mismatch (both >4.3). EdNet-KT1 supports latency and off-task analyses only (V3 0.08; V1 0.52). W1 uses the same log-response-time transform across datasets, but the resulting values remain domain-dependent.

Persona conditioning improved alignment for the three disengaged-state measures shown in Table 2, Panel B. Compared with V3-no-persona, V3-persona reduced the gaming correctness gap from 0.54 to 0.20, the gaming-rate gap from 0.51 to 0.30, and the wheel-spinning correctness gap from 0.51 to 0.18. Post-hoc correctness calibration adjusts the correctness marginal without changing response timing, so it should be interpreted as output calibration rather than evidence of an improved underlying generator. However, closer behavioral alignment did not necessarily produce more distinguishable student states. Five-state classification accuracy was 76.2% for V1 but 30.4% for V3, only moderately above the 20% chance level. V3-cal increased classification accuracy to 45.2%, but its near-zero silhouette score still indicated substantial overlap among states. These results show that behavioral alignment and student-state distinguishability capture different properties of a simulator and should be evaluated separately.

*Table 2. Cross-Dataset Alignment and Persona-Conditioning Ablation*

***Panel A**. Cross-Dataset Response-Time Alignment for the Engaged Student State*

| Dataset | V1 W1 | V3 W1 | Interpretation |
|---|---|---|---|
| ASSISTments09 | 1.00 | 0.50 [0.48, 0.52] | Primary; V3 closer |
| ASSISTments12 | 0.94 | 0.49 | Same ordering |
| Eedi | >4.3 | >4.3 | Domain mismatch |
| EdNet-KT1 | 0.52 | 0.08 | Latency/off-task only |

***Panel B**. Persona-Conditioning Ablation for Disengaged States*

| Measure | No persona | Persona | Reduction |
|---|---|---|---|
| Gaming correctness gap | 0.54 | 0.20 | 0.34 |
| Gaming-rate gap | 0.51 | 0.30 | 0.21 |
| Wheel-spinning correctness gap | 0.51 | 0.18 | 0.33 |

*Note.* Lower values indicate closer alignment. Panel A reports W1 on log1p(latency_ms). Panel B reports absolute simulated-to-authentic gaps for V3. "No persona" uses the engaged-state instruction for every target state, whereas "Persona" uses the matching target-state instruction. EdNet-KT1 lacks per-attempt correctness, so gaming and wheel-spinning are not evaluated.

### *5.3 Tutor Performance Across Student States*

Tutor performance varied across student states, but these differences were small in magnitude. Under the primary GPT-4o-mini evaluator, tutor scores differed significantly across states, but the average changes were approximately 0.01 to 0.03 on the 0 to 1 performance scale. Thus, student state affected the absolute evaluation scores, but the practical magnitude of these differences was limited in the present evaluation. The relative ordering of the five tutors was more stable. Under both the 10-turn and 20-turn evaluations, the tutors appeared in the same order, producing Spearman $\rho = 1.00$. This does not mean that tutors performed identically across student states. Rather, their state-specific scores changed without changing their relative positions. Therefore, a stable tutor ranking can conceal differences in how a tutor performs across student-engagement states. Table 3 summarizes the tutor score and ranking results across evaluation conditions. The 10-turn grid is primary; the 20-turn grid is a robustness check.

*Table 3. Tutor Score and Ranking Results Across Evaluation Conditions*

| Condition | Design | Statistic | Interpretation |
|---|---|---|---|
| 10-turn evaluation (primary) | 5 tutors × 5 states × 40 sessions | $\rho = 1.00$; $\Delta \sim -0.01$ to $-0.02$ | Same ranking, small score changes |
| 20-turn evaluation (robustness) | 5 tutors × 5 states × 40 sessions | $\rho = 1.00$; $\tau = 1.00$ | Same ranking over longer interactions |
| Cross-judge | 200 transcripts | rank $\rho = 0.80$; score $\rho = 0.48$ | Ranking partly depends on evaluator |

The alternative automated judge produced a rank correlation of 0.80 and a score correlation of 0.48 with the primary judge. The highest- and lowest-ranked tutors remained relatively stable, whereas some middle positions changed. This indicates that the overall leaderboard was not fully independent of the automated evaluator used. Human validation further identified limitations in automated tutor evaluation. Human raters showed high agreement with each other, but the original automated evaluator aligned poorly with human judgments of relevance ($\rho = 0.33$). Revising the relevance criterion increased this correlation to 0.61. The resulting overall judge-human correlation reached 0.82, still below the predefined 0.85 benchmark, and the automated evaluator showed slightly greater over-scoring for disengaged than engaged interactions. These results indicate that automated evaluation captured several aspects of tutor quality but was not sufficiently aligned with human judgment to be treated as fully validated.

## 6. Discussion and Conclusion

This study examined whether disengaged student states can be operationalized for student simulation and whether including these states changes the evaluation of AI tutors. Overall, the DAS² student-state definitions were reasonably consistent with human judgments, although gaming and wheel-spinning remained difficult to distinguish. The simulator configurations also revealed a trade-off: closer alignment with authentic interaction distributions did not necessarily produce more distinguishable student states. Finally, tutor scores varied across student states while tutor rankings remained comparatively stable, and the automated evaluator showed incomplete alignment with human judgments. Together, these findings suggest that disengagement-aware student simulation is feasible for pre-deployment evaluation, but behavioral fidelity, state controllability, and tutor evaluation should be treated as separate forms of evidence.

An important implication for student simulation is that reproducing authentic-looking behavior is not sufficient when a simulator is intended to represent a particular student state. Recent work has increasingly evaluated simulated students across multiple behavioral and cognitive dimensions or conditioned them on student histories and knowledge states (Scarlatos et al., 2026; Duan et al., 2026; Song et al., 2026). DAS² extends this direction by showing why state controllability should be evaluated alongside behavioral alignment. For example, a simulator may reproduce realistic response-time distributions while producing gaming, wheel-spinning, and off-task sessions that are difficult to distinguish from one another. For applications that require a specific student condition, such as testing how a tutor responds to persistent failure, realistic aggregate behavior is therefore less useful if the intended state cannot be reliably generated. A practical evaluation of student simulators should consequently ask both whether generated behavior resembles authentic student data and whether the simulator actually represents the student state it was instructed to produce. The findings also have implications for AI tutor evaluation. Gaming, wheel-spinning, and off-task behavior represent different instructional situations and may require different responses, such as redirecting unproductive behavior, providing alternative scaffolding, or re-engaging an inactive student. Evaluating a tutor only with engaged or cooperative simulated students therefore provides evidence about a relatively limited interaction condition. Moreover, a stable leaderboard should not be interpreted as evidence that tutors respond equally well across student states. Tutor rankings can remain unchanged even when individual tutors respond differently to disengaged students. This complements prior work showing that aggregate educational-system performance can conceal meaningful differences experienced by particular groups of students (Doroudi & Brunskill, 2019). For pre-deployment evaluation, state-specific performance therefore provides information that an overall score or ranking alone cannot capture.

Several limitations should be noted. Human validation used 100 ASSISTments09 sessions, with limited off-task cases and some ambiguity between gaming and wheel-spinning. The operational rules capture selected forms of disengagement rather than the full engagement construct. V2 does not fully separate retrieval and evaluation data, and V3-cal adjusts correctness through post-processing. Because V1 and V3 use different underlying models, their differences cannot be attributed solely to fine-tuning. Tutor evaluation was also limited to five models and relied mainly on one automated judge, while simulated interactions do not measure actual student learning. Future work should validate DAS² across larger and more diverse datasets, use stricter held-out evaluation and controlled simulator comparisons, and examine whether disengagement-aware tutor performance predicts human judgments and real student outcomes.

**Reproducibility Statement**

Supplementary materials are available in the project GitHub repository.[1]

## Acknowledgements

This work was supported by the Research Grants Council (Hong Kong) under Early Career Scheme, Grant/Award Number: #27608026.

---

[1] https://github.com/GEMLabHKU/DAS---supplement/blob/main/DAS2_ICCE2026_Appendix.md